\documentclass{SCIS2021}
\usepackage{array, multirow}
\usepackage{cite}
\begin{document}
%\oa
%%%%%%%%%%%%%%%%%%%%%%%%%%%%%%%%%%%%%%%%%%%%%%%%%%%%%%%
%%% Authors do not modify the information below
%%% 作者不需要修改此处信息
\ArticleType{RESEARCH PAPER}
%\SpecialTopic{}
%\luntan
\Year{2020}
\Month{}
\Vol{}
\No{}
\DOI{}
\ArtNo{}
\ReceiveDate{}
\ReviseDate{}
\AcceptDate{}
\OnlineDate{}
%%%%%%%%%%%%%%%%%%%%%%%%%%%%%%%%%%%%%%%%%%%%%%%%%%%%%%%

%%% title: 标题
%%%   \title{title}{title for citation}
\title{FAF: A Feature-Adaptive Framework for Few-Shot Time Series Forecasting}

%%% Corresponding author: 通信作者
%%%   \author[number]{Full name}{{email@xxx.com}}
%%% General author: 一般作者
%%%   \author[number]{Full name}{}
\author[1]{Pengpeng OUYANG}{}
\author[1,2,3,4]{Dong CHEN}{chendongai@zzu.edu.cn}
\author[1]{Tong YANG}{}
\author[1,2,3,4]{Shuo FENG}{}
\author[1,2,3,4]{\texorpdfstring{\\}{}Zhao JIN}{}
\author[1,2,3,4]{Mingliang XU}{iexumingliang@zzu.edu.cn}

%%% Author information for page head. 页眉中的作者信息
\AuthorMark{Pengpeng OUYANG}

%%% Authors for citation. 首页引用中的作者信息
\AuthorCitation{Pengpeng OUYANNG, Dong CHEN, Tong YANG, et al}

%%% Authors' contribution. 同等贡献
%\contributions{Authors A and B have the same contribution to this work.}

%%% Address. 地址
%%%   \address[number]{Affiliation, City {\rm Postcode}, Country}

\address[1]{The School of Computer and Artificial Intelligence of Zhengzhou University, Zhengzhou 450001, China}
\address[2]{Engineering Research Center of Intelligent Swarm Systems, Ministry of Education, Zhengzhou 450001, China}
\address[3]{National Supercomputing Center In Zhengzhou, Zhengzhou 450001, China}
\address[4]{Henan Large model Technology and New quality Software Engineering Research Center, Zhengzhou 450001, China}

%%% Abstract. 摘要
\abstract{
Multi-task and few-shot time series forecasting tasks are commonly encountered in scenarios such as the launch of new products in different cities. However, traditional time series forecasting methods suffer from insufficient historical data, which stems from a disregard for the generalized and specific features among different tasks. For the aforementioned challenges, we propose the Feature-Adaptive Time Series Forecasting Framework (FAF), which consists of three key components: the Generalized Knowledge Module (GKM), the Task-Specific Module (TSM), and the Rank Module (RM).
During training phase, the GKM is updated through a meta-learning mechanism that enables the model to extract generalized features across related tasks. Meanwhile, the TSM is trained to capture diverse local dynamics through multiple functional regions, each of which learns specific features from individual tasks. During testing phase, the RM dynamically selects the most relevant functional region from the TSM based on input sequence features, which is then combined with the generalized knowledge learned by the GKM to generate accurate forecasts. 
This design enables FAF to achieve robust and personalized forecasting even with sparse historical observations
We evaluate FAF on five diverse real-world datasets under few-shot time series forecasting settings. Experimental results demonstrate that FAF consistently outperforms baselines that include three categories of time series forecasting methods. 
In particular, FAF achieves a 41.81\% improvement over the best baseline, iTransformer, on the CO$_2$ emissions dataset.
}

%%% Keywords. 关键词

\maketitle

%%%%%%%%%%%%%%%%%%%%%%%%%%%%%%%%%%%%%%%%%%%%%%%%%%%%%%%
%%% The main text. 正文部分
%%%%%%%%%%%%%%%%%%%%%%%%%%%%%%%%%%%%%%%%%%%%%%%%%%%%%%%
%%%%%%%%%%%%%%%%%%%%%%%%%%%%%%%%%%%%%%%%%%%%%%%%%%%%%%%
%%% 第一节
%%%%%%%%%%%%%%%%%%%%%%%%%%%%%%%%%%%%%%%%%%%%%%%%%%%%%%%
\section{Introduction}\label{sec:intro}
Time series forecasting is a important research area in data mining and machine learning~\cite{1}, which plays an essential role in decision-making in various fields, including finance~\cite{2}, meteorology~\cite{3}, energy management~\cite{4}, and transportation systems~\cite{5}.

\begin{figure}[!t]
\centering
\includegraphics[width=0.5\textwidth]{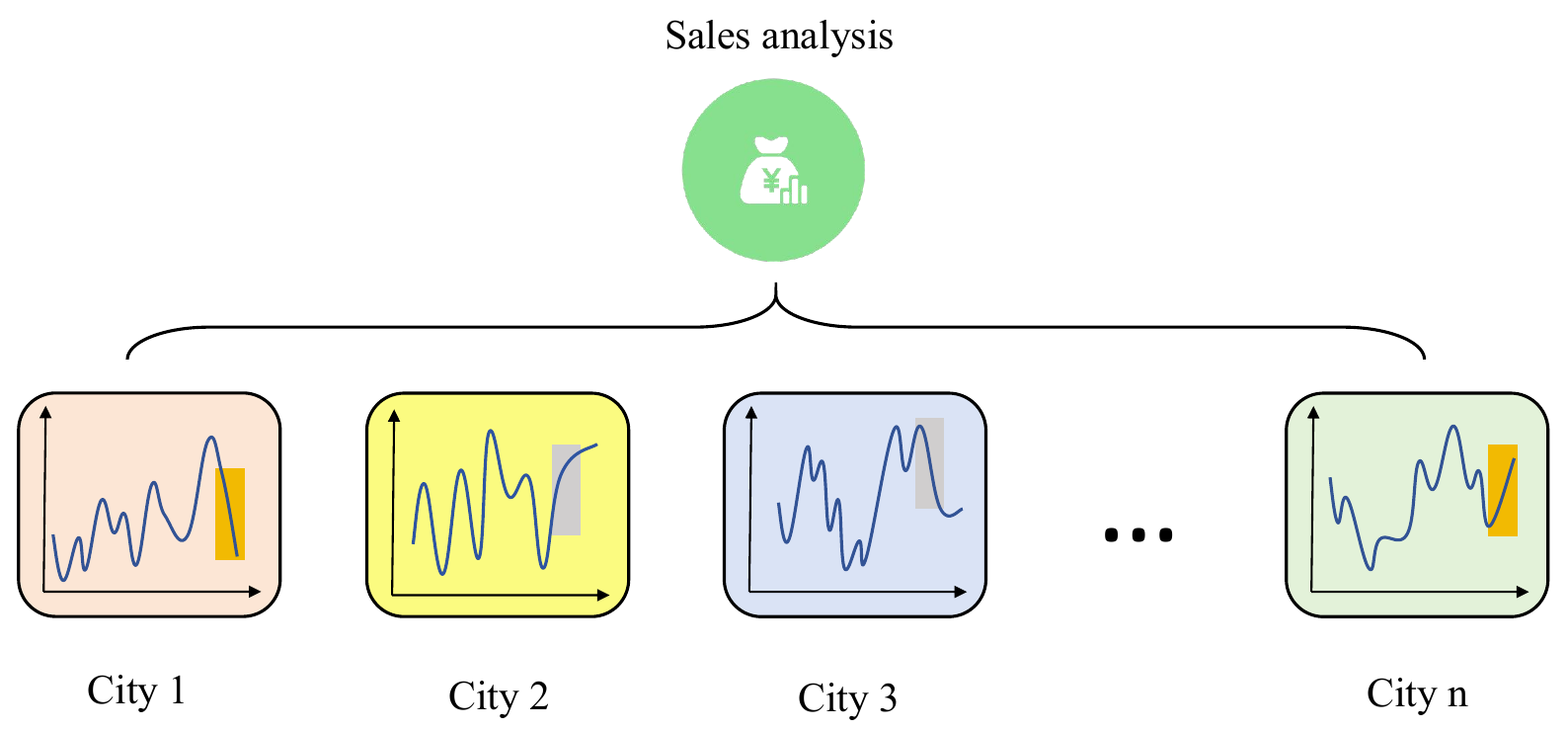}
\caption{Daily sales time series of a newly launched product across multiple cities}
\label{fig1}
\end{figure}

Existing time series forecasting methods have achieved remarkable performance in relevant tasks by capturing complex temporal dependencies within extensive historical sequences~\cite{6,7,8,9}. However, these methods frequently demonstrate limited generalization capabilities in few-shot scenarios due to overfitting and gradient inconsistency~\cite{10}. As shown in Table~\ref{tab1}, traditional time series forecasting methods such as ARIMA~\cite{11}, Prophet~\cite{12}, and SARIMA~\cite{13} produce high prediction errors across RMSE, MAE, and MAPE metrics. This performance indicates their limited effectiveness when only limited historical sequences are available in cold-start or few-shot scenarios. Meta-learning is one of the most important keys for cold-start and few-shot scenarios, which can improve the performance of time series forecasting models by learning generalization features across diverse tasks~\cite{14,15,16,17}. These generalization features enhance the model’s ability of capturing overall trend patterns across time series forecasting tasks, thereby providing a robust foundation for few-shot time series forecasting. In contrast, task-specific features offer fine-grained complementary knowledge that improves predictive accuracy by adapting to local variations. More specifically, we define generalization features as global trend patterns shared across related tasks. As illustrated in Figure~\ref{fig1}, these patterns often represent overall temporal trends, such as the consumption cycle of a newly launched product. As for task-specific features, we define them as local variations unique to individual time series. As shown in Figure~\ref{fig1}, each city exhibits distinct market behaviors, including regional consumption habits and seasonal fluctuations. Both generalization features and task-specific features play crucial roles in time series forecasting, contributing to overall trend estimation and localized adaptation, respectively.

\begin{table}[!t]
\centering
\caption{Performance of ARIMA, SARIMA, and Prophet on limited historical sequences for electricity load forecasting.}
\label{tab1}
\tabcolsep 25pt %space between two columns. 用于调整列间距
\begin{tabular}{lccc}
\toprule
Method & RMSE & MAE & MAPE (\%) \\
\midrule
ARIMA   & 769.0921 & 699.7889 & 98.4241 \\
SARIMA  & 816.4226 & 738.5723 & 99.7622 \\
Prophet & 743.8059 & 685.7275 & 98.1015 \\
\bottomrule
\end{tabular}
\end{table}

\begin{figure}[!t]
\centering
\includegraphics[width=1.0\textwidth]{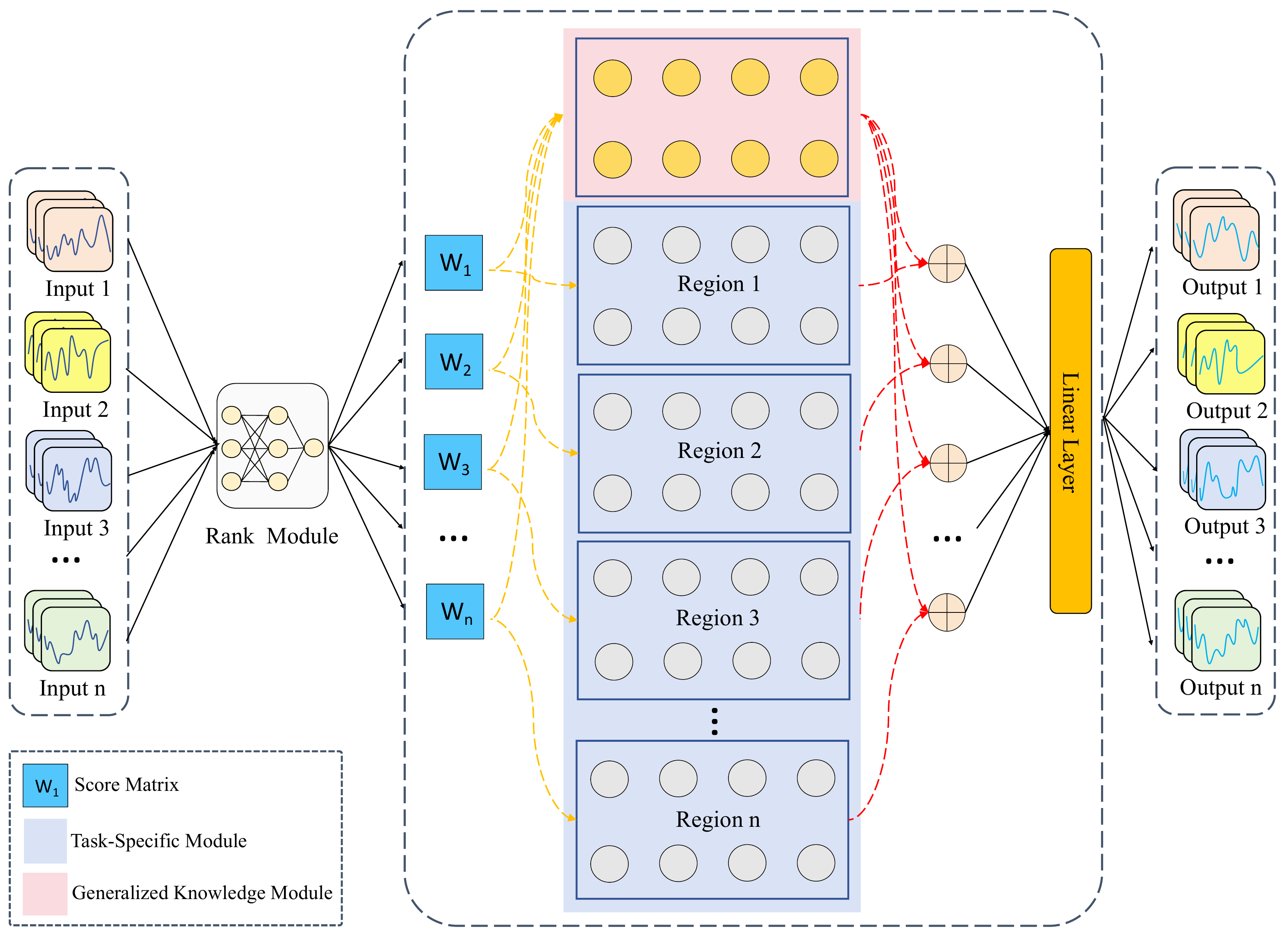}
\caption{Overview of the FAF framework for few-shot time series forecasting.}
\label{fig2}
\end{figure}

Although models based on neural networks are inherently capable of learning both high-level generalization features and low-level task-specific features~\cite{18}, the conflict among task-specific features across different tasks can degrade the overall learning performance. This is due to conflicting trend dynamics among different time series, where parts of local inconsistent temporal trends across tasks hinder model grasping local inherent trends. For instance, as illustrated in Figure~\ref{fig1}, the yellow-shaded regions in City 1 and City n exhibit trend-inconsistent behaviors, while the gray-shaded areas in City 2 and City 3 highlight local variations that may interfere with the extraction of generalization features. These feature conflicts hinder model adaptability and degrade forecasting accuracy under data-scarce conditions. To effectively utilize cross-task generalization features in capturing global temporal patterns while mitigating conflicts among task-specific features, we propose the  Feature-Adaptive Time Series Forecasting Framework (FAF). As illustrated in Figure~\ref{fig2}, FAF consists of three key components: the Generalized Knowledge Module (GKM), the Task-Specific Module (TSM), and the Rank Module (RM). 
During the traning phase, the GKM learns generalization features through a meta-learning mechanism. These features are derived from gradient updates of related tasks during multi-task training and enable the module to rapidly adapt to new time series with limited historical sequences ~\cite{12}. As for the TSM, as shown in the blue segments of Figure~\ref{fig2}, we divide multiple function regions, each specializing in a unique type of local temporal dynamics to learn different task-specific features. The RM is trained end-to-end with the rest of the FAF, learning to assign appropriate weights to each region in the TSM based on its relevance to the target time series.
During the inference phase, FAF dynamically selects the most relevant function region within the TSM based on the RM, which will evaluate the similarity between different input sequences and function regions to identify the best-matching local pattern. Simultaneously, the GKM processes the entire input sequence to extract generalized features, and the output will be combined with the output of the selected function regions in the TSM to produce accurate time series prediction results. 

The main contributions of this work are summarized as follows:
\begin{itemize}
\item We analyze the limitations of existing time series forecasting methods in few-shot scenarios and discuss the underlying causes from the perspective of generalization features and task-specific features;
\item We propose the FAF framework, which consists of GKM, TSM and RM, effectively alleviating feature conflicts and improving the accuracy of time series forecasting in few-shot scenarios;
\item We evaluate the effectiveness of the proposed framework on various practical dataset. In particular, FAF achieves a 41.81\% improvement over the best baseline, iTransformer, on the CO$_2$ emissions dataset.
\end{itemize}

%%%%%%%%%%%%%%%%%%%%%%%%%%%%%%%%%%%%%%%%%%%%%%%%%%%%%%%
%%% 相关工作与总结
%%%%%%%%%%%%%%%%%%%%%%%%%%%%%%%%%%%%%%%%%%%%%%%%%%%%%%%
\section{Related work}

\subsection{Time series forecasting}

Time series forecasting is fundamentally the process of estimating future values through statistical analysis of sequential historical observations~\cite{13}. Traditional statistical models such as ARIMA, Prophet, and SARIMA rely on restrictive assumptions of linearity, stationarity, and seasonality, limiting their adaptability to complex real-world dynamics~\cite{19}. Subsequently developed deep learning approaches including recurrent neural networks (RNNs)~\cite{20}, long short-term memory networks (LSTMs)~\cite{21}, and attention-based architectures like the Transformer~\cite{22}, have significantly advanced prediction capabilities by capturing nonlinear dependencies and long-range temporal patterns. However, these methods generally require substantial historical data for effective training. Under limited-data conditions, they remain susceptible to overfitting and unstable gradient updates, resulting in performance degradation in cold-start or few-shot scenarios. These limitations have consequently accelerated research into specialized few-shot learning paradigms for time series forecasting.

\subsection{Few-shot learning in time series forecasting}
Few-shot learning aims to enable models to generalize effectively from limited data~\cite{23}. 
Although the emergence of large language models has introduced significant progress in this field~\cite{24,25,26,27}, existing studies suggest that their generalization capabilities remain limited in the context of time series forecasting~\cite{28}. One widely adopted approach to address this challenge is meta-learning, which enables models to acquire transferable knowledge across related tasks and facilitates rapid adaptation to new tasks using only a few samples~\cite{14}. In time series forecasting, meta-learning is commonly used to extract generalization features that capture shared global trend patterns across multiple sequences~\cite{29}.These features provide a robust initialization for fast adaptation and have demonstrated strong performance in improving model generalization under scarce-data conditions. 
Although meta-learning has demonstrated effectiveness in few-shot time series forecasting, most existing approaches primarily focus on generalization features and often overlook the importance of task-specific features. Task-specific features capture fine-grained local variations unique to individual time series and are essential for accurate localized adaptation. However, when conflicting local variations exist across different tasks, these heterogeneous patterns may interfere with each other, degrading model performance. Such feature conflicts hinder adaptability and reduce forecasting accuracy, especially in data-scarce scenarios.
In contrast, our method explicitly decouples generalization and task-specific features, addressing feature conflicts through a dynamic adaptation mechanism.

%%%%%%%%%%%%%%%%%%%%%%%%%%%%%%%%%%%%%%%%%%%%%%%%%%%%%%%
%%% 第二节
%%%%%%%%%%%%%%%%%%%%%%%%%%%%%%%%%%%%%%%%%%%%%%%%%%%%%%%
\section{Methodology}
As illustrated in Figure~\ref{fig2}, the FAF framework adopts a modular architecture that learns generalization and task-specific features in parallel. Given an input time series $ x \in \mathbb{R}^{L_{\text{in}}} $, the Generalized Knowledge Module first extracts its global trend pattern using shared parameters across all tasks. Meanwhile, the Rank Module evaluates the compatibility between the input and each functional region within the Task-Specific Module, selecting the most relevant ones based on feature similarity. The final prediction is obtained by adaptively fusing the outputs from these selected regions and Generalized Knowledge Module, ensuring both robust trend modeling and fine-grained adaptation to local dynamics.

This section begins with a detailed description of the FAF architecture. Following that, we discuss the meta-learning based training process.
\subsection{Model architecture}
The FAF framework integrates three core components enabling feature-adaptive few-shot time series forecasting: (1) the Generalized Knowledge Module, which extracts general trend across tasks, (2) the Task-Specific Modules dedicated to capturing localized sequence variations, and (3) the Rank Module that dynamically selects relevant functional regions within Task-Specific Modules during inference phase.

\textbf{Generalized Knowledge Module.}
The Generalized Knowledge Module extracts generalization features across all tasks and serves as the foundation for fast adaptation. It is implemented as a multi-layer perceptron:
\begin{equation}
\label{eq3}   
z = f_{\text{generalized}}(x; W_g)
\end{equation}
where $ x \in \mathbb{R}^{L_{\text{in}}} $ denotes the input time series of length $ L_{\text{in}} $. The output $ z \in \mathbb{R}^{L_{\text{in}}} $ represents the extracted generalization features. The parameter matrix $ W_g $ governs the transformation applied to the input sequence. The function $ f_{\text{generalized}}(\cdot) $ applies non-linear transformations to extract high-level trend features that are transferable across different tasks.

During training, this module undergoes meta-learning updates using support set gradients, ensuring stable initialization even when only limited history is available for new tasks.

\textbf{Task-Specific Module.}
The Task-Specific Module contains multiple functional regions, each responsible for modeling distinct local variation patterns. Each functional region is defined as an independent neural network with parameter matrix $ W_s^{(j)} $, where $ j \in [1, N] $ indexes the $ j $-th functional region.
\begin{equation}
\label{eq4}   
y_j = f_{\text{task-specific}}^{(j)}(x; W_s^{(j)})
\end{equation}
Here, $ x \in \mathbb{R}^{L_{\text{in}}} $ is the input time series of length $ L_{\text{in}} $, and $ y_j \in \mathbb{R}^{L_{\text{out}}} $ is the output prediction generated by the $ j $-th functional region. The parameter matrix $ W_s^{(j)} $ determines how the selected functional region processes its input to capture local fluctuations specific to task $ i $, and $ f_{\text{task-specific}}^{(j)}(\cdot) $ denotes its forward function.

During inference, only a subset of these regions is activated based on the Rank Module. Then, their outputs are weighted and combined with the output from the Generalized Knowledge Module to generate the final forecast.

\textbf{Rank Module.}
To enable dynamic module selection, we introduce the Rank Module, which computes relevance scores between the input and each functional region within the Task-Specific Module. Specifically, this module performs linear projection followed by normalization via softmax.
\begin{equation}
\label{eq5}   
s(x) = W_r x + b,
\quad p_j = \frac{\exp(s_j(x))}{\sum_{j'} \exp(s_{j'}(x))}
\end{equation}
In this formulation, $ x \in \mathbb{R}^{L_{\text{in}}} $ is the input time series, and $ s(x) \in \mathbb{R}^N $ is the raw score vector over all functional regions. Each element $ s_j(x) $ measures the matching degree between the input and the $ j $-th region. After applying the softmax function, the resulting value $ p_j \in [0, 1] $ indicates the normalized probability that the $ j $-th region should be activated. The weight matrix $ W_r \in \mathbb{R}^{N \times L_{\text{in}}} $ and bias term $ b \in \mathbb{R}^N $ are learnable parameters used in this ranking process.

During inference, the top-$k$ most relevant regions are selected based on these probabilities.
\begin{equation}
\label{eq6}   
\mathcal{E}_{\text{active}}^{(i)} = \text{Top-}k(p_1^{(i)}, p_2^{(i)}, ..., p_N^{(i)}), \quad \text{for task } i.
\end{equation}
This mechanism ensures that the model can dynamically compose forecasting strategies according to the characteristics of the input sequence.

\textbf{Dynamic Knowledge Composition.}
in inference phase, the final prediction is computed by combining knowledge from both the Generalized Knowledge Module and the selected functional regions within Task-Specific Module:
\begin{equation}
\label{eq7}   
\hat{y}^{(i)} = f_{\text{fusion}}\left(f_{\text{generalized}}(x; W_g') + \sum_{j \in \mathcal{E}_{\text{active}}^{(i)}} w_j^{(i)} f_{\text{task-specific}}^{(j)}(x; W_s^{(j)})\right)
\end{equation}
In this formulation, $ \hat{y}^{(i)} \in \mathbb{R}^{L_{\text{out}}} $ represents the final prediction for task $ i $. The updated version of the Generalized Knowledge Module after fast adaptation is denoted by $ W_g' $. The weight $ w_j^{(i)} $ reflects the contribution of the $ j $-th functional region to the prediction of task $ i $. The function $ f_{\text{fusion}}(\cdot) $ is the regression layer responsible for mapping the aggregated features into the final output space.

\subsection{Model Training}
The FAF framework is trained using a meta-learning-inspired training strategy that enables rapid adaptation to new tasks with limited historical data. Unlike conventional multi-task learning approaches, our training process explicitly separates generalizable knowledge from task-specific features and updates them in distinct ways.

During each training iteration, we sample a batch of tasks from the training set, where each task corresponds to a specific entity such as a city or store. For each sampled task $ i $, we use its full historical sequence to compute gradients for both the Generalized Knowledge Module and the corresponding Task-Specific Module. This is in contrast to the validation and testing phases, where only a subset of historical data, referred to as the \textit{support set}, is used for fast parameter adaptation, while the remaining data constitutes the \textit{query set} for performance evaluation.

The Generalized Knowledge Module accumulates gradients across all sampled tasks during each training iteration. Specifically, after computing individual gradients from each task, we perform gradient averaging to update the shared parameters.
\begin{equation}
\label{eq8}
W_g \leftarrow W_g - \beta \cdot \frac{1}{B} \sum_{i=1}^{B} \nabla_{W_g} \mathcal{L}_{\text{task}}^{(i)}(f_{\text{generalized}}(x^{(i)}; W_g))
\end{equation}
where $ B $ is the number of tasks sampled per iteration, $ \beta $ is the meta-learning rate, and $ \mathcal{L}_{\text{task}}^{(i)} $ represents the loss function evaluated on task $ i $.

On the other hand, each functional regional within Task-Specific Module is updated exclusively with the gradients derived from its associated task.
\begin{equation}
\label{eq9}
W_s^{(i)} \leftarrow W_s^{(i)} - \alpha \cdot \nabla_{W_s^{(i)}} \mathcal{L}_{\text{task}}^{(i)}(f_{\text{task-specific}}^{(i)}(x^{(i)}; W_s^{(i)}))
\end{equation}
where $ W_s^{(i)} $ are the parameters of the functional regions associated with task $ i $ and $ \alpha $ is the task-specific learning rate.

To further evaluate the model's performance when limited historical data is available, we simulate few-shot scenarios during test set. In the setting, we construct a support set $ D_i^{\text{supp}} $ and a query set $ D_i^{\text{query}} $ for each task $ i $. The support set is used to perform a single-step gradient update on the Generalized Knowledge Module.
\begin{equation}
\label{eq10}
W_g' = W_g - \alpha \cdot \nabla_{W_g} \mathcal{L}_{\text{supp}}^{(i)}(f_{\text{generalized}}(x^{(i)}; W_g))
\end{equation}
where $ \mathcal{L}_{\text{supp}}^{(i)} $ evaluates the model’s performance on the support set. The updated module $ W_g' $ is then used to generate predictions on the query set $ D_i^{\text{query}} $, allowing us to assess the model's accuracy of time series forecasting in few-shot scenario.

Finally, the Rank Module is trained end-to-end alongside the rest of the model. It learns to assign weights $ w_j^{(i)} $ to the outputs of the selected functional regions within Task-Specific Module based on input feature similarity. These weights determine how the final prediction is composed, enabling the model to dynamically balance generalization and task-specific features in inference phase.
\begin{equation}
\label{eq11}
W_r \leftarrow W_r - \alpha \cdot \nabla_{W_r} \mathcal{L}_{\text{total}}\left(f_{\text{fusion}}\left(f_{\text{generalized}}(x; W_g') + \sum_{j \in \mathcal{E}_{\text{active}}^{(i)}} w_j^{(i)} f_{\text{task-specific}}^{(j)}(x; W_s^{(j)})\right), y \right)
\end{equation}

In addition, we use a joint loss function that integrates mean squared error with load-balancing regularization.
\begin{equation}
\label{eq12}   
\mathcal{L}_{\text{total}} = \mathcal{L}_{\text{MSE}} + \lambda \cdot \mathcal{L}_{\text{balance}}
\end{equation}
where $ \mathcal{L}_{\text{MSE}} $ quantifies the prediction error against ground truth values, $\lambda$ is the balancing factor that regulates the contribution of the load-balancing regularization term, and $\mathcal{L}_{\text{balance}} $ encourages fair usage of functional regions within the Task-Specific Module.
\begin{equation}
\label{eq13}   
\mathcal{L}_{\text{balance}} = \sum_{j=1}^{N} \left( \text{count}_j - \bar{\text{count}} \right)^2
\end{equation}
where $ \text{count}_j $ denotes how often region $ j $ is activated across all tasks, and $ \bar{\text{count}} $ represents the average activation count. This term penalizes the functional regions in Task-Specific Module that are either overly utilized or rarely selected, promoting diversity in the usage of regions.

%%%%%%%%%%%%%%%%%%%%%%%%%%%%%%%%%%%%%%%%%%%%%%%%%%%%%%%
%%% 第三节
%%%%%%%%%%%%%%%%%%%%%%%%%%%%%%%%%%%%%%%%%%%%%%%%%%%%%%%
\section{Dataset}
\subsection{Dataset discription}
In this section, we provide a brief introduction of the five real-world time series forecasting datasets used in our experiments. These datasets span diverse application domains, including electricity load management, retail sales prediction, carbon emissions analysis, macroeconomic growth modeling, and temperature monitoring.

The detailed descriptions of each dataset are as follows:

(1) Electricity Load Dataset \footnote{Available online: \url{https://www.iso-ne.com/isoexpress/web/reports/load-and-demand}}:  This dataset contains monthly electricity consumption records from eight load zones in New England, spanning from January 2016 to December 2024.

(2) Walmart Sales Dataset \footnote{Available online: \url{https://www.kaggle.com/datasets/varsharam/walmart-sales-dataset-of-45stores}}:  The Walmart Sales Dataset consists of weekly sales data from 45 U.S. stores between 2010 and 2012, capturing regional purchasing behaviors across locations.

(3) CO$_2$ Emissions Dataset \footnote{Available online: \url{https://www.kaggle.com/datasets/saloni1712/co2-emissions}}:  This dataset includes daily CO$_2$ emission records from 84 departments across 14 countries, covering the period from January 1, 2019 to May 31, 2023.

(4) GDP Growth Dataset \footnote{Available online: \url{https://www.kaggle.com/datasets/stealthtechnologies/gdp-growth-of-european-countries}}:  The GDP Growth Dataset collects annual economic growth rates from 18 European countries, ranging from 1960 to 2023.

(5) Temperature Forecasting Dataset \footnote{Available online: \url{https://www.kaggle.com/datasets/vanvalkenberg/historicalweatherdataforindiancities}}:  This dataset documents daily average temperature readings from eight Indian cities, recorded between January 1, 1990 and July 20, 2022.

\subsection{Dataset preprocessing}

We preprocess all datasets through a standardized pipeline comprising four stages: task partitioning, feature normalization, input-output window construction, and support/query set splitting. First, each dataset is divided into independent time series forecasting tasks based on geographic regions (e.g., cities), store IDs, or country-level indicators. Specifically, the Electricity Load dataset is organized by cities, the Walmart Sales dataset by stores, the CO$_2$ Emissions dataset by departments within countries, the GDP Growth dataset by European countries, and the Temperature Forecasting dataset by Indian cities. Subsequently, Tasks in each dataset are split into training (80\%), validation (10\%), and test (10\%) sets, with detailed statistics presented in Table~\ref{tab2} using the format "Tasks/Time Steps". In this format, the value before the slash indicates the number of distinct tasks assigned to the subset, while the value after the slash represents the total number of time steps available per task.

As shown in Equation~\ref{eq1}, we apply feature standardization using statistics computed from the training set.
\begin{equation}
\label{eq1}   
x' = \frac{x - \mu_{\text{train}}}{\sigma_{\text{train}} + \epsilon}
\end{equation}
where $ x $ denotes the original feature, $ \mu_{\text{train}} $ and $ \sigma_{\text{train}} $ represent the mean and standard deviation calculated from the training set, and $ \epsilon $ is a small constant to prevent division by zero. This transformation stabilizes gradient propagation during training while preserving local variations across different tasks.

\begin{table}[!t]
\centering
\caption{Dataset Statistics.}
\label{tab2}
\tabcolsep 25pt %space between two columns. 用于调整列间距
\begin{tabular}{lccc}
\toprule
Dataset & Train Set & Validation Set & Test Set \\
\midrule
Electricity  & 6 / 140 & 1 / 140 & 1 / 24 \\
Walmart    & 36 / 140 & 4 / 140 & 5 / 24 \\
CO$_2$    & 67 / 140 & 8 / 140 & 9 / 24 \\
GDP      & 14 / 140 & 2 / 140 & 2 / 24 \\
Temperature       & 6 / 140 & 1 / 140 & 1 / 24 \\
\bottomrule
\end{tabular}
\end{table}

To further increase sample diversity and support better convergence during meta-training we construct input-output pairs using a sliding window approach. Given a time series $ S = \{x_1 x_2 \dots x_T\} $ we generate multiple historical instances by sliding a fixed-size window over the sequence. As shown in Equation~\ref{eq2}, we define the input window size $ L_{in} = 3 $ and the output window size $ L_{out} = 1 $. 
\begin{equation}
\label{eq2}   
X^{(i)} = [x_i, x_{i+1}, x_{i+2}], \quad Y^{(i)} = [x_{i+3}]
\end{equation}
This method allows the model to learn from diverse historical segments and improves gradient estimation during parameter updates.

Finally, few-shot time series forecasting is simulated by limiting adaptation historical sequence. For test tasks, only 16/24 time steps form the support set for rapid adaptation, while the remainder constitute the query set. This setup closely resembles the real-world few-shot scenario where new tasks may have minimal prior observations to learn.

%%%%%%%%%%%%%%%%%%%%%%%%%%%%%%%%%%%%%%%%%%%%%%%%%%%%%%%
%%% 第四节
%%%%%%%%%%%%%%%%%%%%%%%%%%%%%%%%%%%%%%%%%%%%%%%%%%%%%%%
\section{Experiments}
\subsection{Experimental Setup}
\textbf{Evaluation metrics.} 
To evaluate the performance of the model, we use three metrics: The root mean square error (RMSE), the mean absolute error (MAE), and the mean absolute percentage error (MAPE). Lower values in all of these metrics indicate superior forecast accuracy. Definitions of these metrics are:
\begin{equation}
\label{eq14}   
RMSE = \sqrt{\frac{1}{n}\sum_{i=1}^{n}(y_i - \hat{y}_i)^2}
\end{equation}
\begin{equation}
\label{eq15}   
MAE = \frac{1}{n}\sum_{i=1}^{n}|y_i - \hat{y}_i|
\end{equation}
\begin{equation}
\label{eq16}   
MAPE = \frac{100}{n}\sum_{i=1}^{n}\left|\frac{y_i - \hat{y}_i}{y_i}\right|
\end{equation}

\textbf{Baseline Methods.} 
To comprehensively demonstrate the effectiveness of our proposed method, we compare FAF with three categories of time series forecasting approaches. The baseline methods include (1) classical deep learning models for time series forecasting, including LSTM~\cite{21}, BiLSTM~\cite{30}, CNN-LSTM~\cite{31} and GRU~\cite{32}; (2)transformer and time series transformers, including Transformer~\cite{22}, PatchTST~\cite{33}, Autoformer~\cite{34}, Pyraformer~\cite{35}, Informer~\cite{36} and iTransformer~\cite{37}; (3) Other popular methods, including TimesNet~\cite{38}, Nlinear~\cite{39}, DLinear~\cite{39} and NBeats~\cite{40}.

\textbf{Implementation Details.} 
All experiments are conducted on an NVIDIA RTX 4090 GPU with 24GB memory. We adopt a consistent experimental setup across all datasets, where each test task serves as a target task under the few-shot forecasting setting. The historical time series length is fixed at 16 time steps for fast adaptation, and the model predicts the subsequent 8 time steps. Final results are reported as the average performance over all test tasks.
During training, we use a batch size of 4 and optimize the model using the Adam optimizer. The meta-learning rate $\beta$, which governs the update of the Generalized Knowledge Module, is set to $1 \times 10^{-5}$. The task-specific learning rate $\alpha$, used for updating the Rank Module and Task-Specific Module, is set to $1 \times 10^{-3}$. Additionally, we introduce a balancing factor $\lambda = 1 \times 10^{-5}$ to encourage fair utilization of functional regions within the Task-Specific Module, preventing certain regions from being overly favored or neglected.
To ensure stable training and prevent overfitting, we apply gradient clipping with a maximum norm of 1.0 and employ early stopping based on validation performance. Input features are normalized using Z-score normalization based on training set statistics. All model parameters are initialized using PyTorch's default initialization scheme and trained until convergence, with the best-performing checkpoint selected based on validation loss.

\begin{table}[!t]
\centering
\caption{Performance comparisons on five real-world datasets in terms of RMSE, MAE, and MAPE.The best is in bold, while the second best is underlined.}
\label{tab3}
\resizebox{\textwidth}{!}{
\begin{tabular}{l|ccc|ccc|ccc|ccc|ccc}
\toprule
{Dataset} & \multicolumn{3}{c|}{Electricity} & \multicolumn{3}{c|}{Walmart} & \multicolumn{3}{c|}{CO$_2$} & \multicolumn{3}{c|}{GDP} & \multicolumn{3}{c}{Temperature} \\
\cline{1-16}
Metric & RMSE & MAE & MAPE (\%) & RMSE & MAE & MAPE (\%) & RMSE & MAE & MAPE (\%) & RMSE & MAE & MAPE (\%) & RMSE & MAE & MAPE (\%) \\
\midrule
\textbf{FAF (ours)} & \underline{0.1251} & \textbf{0.1021} & \textbf{10.0156} & \textbf{0.0529} & \textbf{0.0450} & \textbf{6.8016} & \textbf{0.0060} & \textbf{0.0055} & \textbf{1.5923} & \textbf{0.0083} & \textbf{0.0073} & \textbf{1.0997} & 0.5917 & 0.5059 & \textbf{17.4309} \\
LSTM & 0.1288 & 0.1062 & 10.5633 & 0.0544 & 0.0469 & 8.3335 & 0.0215 & 0.0212 & 5.9959 & 0.0801 & 0.0800 & 12.0585 & 0.7039 & 0.5708 & 20.7839 \\
BiLSTM & 0.1258 & 0.1058 & 10.2257 & 0.0552 & 0.0468 & \underline{7.1134} & 0.0120 & 0.0103 & 2.7858 & 0.0239 & 0.0198 & 2.9602 & 0.6489 & 0.5583 & 19.8188 \\
CNN-LSTM & 0.2453 & 0.2029 & 18.7556 & 0.1922 & 0.1855 & 26.7971 & 0.0196 & 0.0187 & 5.3687 & 0.0194 & 0.0188 & 2.9066 & 0.7784 & 0.6959 & 24.2793 \\
GRU & 0.2115 & 0.1796 & 17.0635 & 0.1280 & 0.1126 & 15.1987 & 0.0338 & 0.0287 & 8.0012 & 0.0619 & 0.0433 & 6.5703 & 0.6892 & 0.5705 & 20.2650 \\
\midrule % Add horizontal line after GRU
Transformer & 0.2297 & 0.1871 & 17.6235 & 0.0924 & 0.0801 & 16.0699 & 0.0681 & 0.0534 & 14.6392 & 0.1049 & 0.0852 & 12.9126 & \underline{0.5386} & \underline{0.4598} & 17.8876 \\
Autoformer & 0.1218 & \underline{0.1024} & \underline{10.0232} & 0.1053 & 0.0825 & 12.5237 & 0.0364 & 0.0308 & 8.3092 & 0.1082 & 0.0919 & 13.9712 & \textbf{0.5332} & \textbf{0.4584} & \underline{17.5299} \\
Informer & 0.2324 & 0.2057 & 18.9505 & 0.1643 & 0.1492 & 20.5483 & 0.0946 & 0.0789 & 21.7686 & 0.1496 & 0.1222 & 19.4717 & 0.9454 & 0.8238 & 28.9410 \\
PatchTST & 0.1912 & 0.1571 & 15.1097 & 0.1095 & 0.0874 & 16.3444 & 0.0427 & 0.0358 & 9.9411 & 0.1259 & 0.1050 & 15.7408 & 0.5458 & 0.4653 & 17.9739 \\
iTransformer & 0.1340 & 0.1163 & 11.6500 & 0.0556 & 0.0461 & 7.1206 & \underline{0.0089} & \underline{0.0078} & \underline{2.1589} & \underline{0.0105} & \underline{0.0094} & \underline{1.4005} & 0.6395 & 0.5530 & 19.9428 \\
Pyraformer & 0.1515 & 0.1253 & 12.7185 & 0.0582 & 0.0481 & 7.2133 & 0.0138 & 0.0134 & 3.7422 & 0.0325 & 0.0320 & 4.8085 & 0.6569 & 0.5529 & 19.6956 \\
\midrule % Add horizontal line after Pyraformer
DLinear & 0.1248 & 0.1066 & 10.4159 & 0.0554 & 0.0479 & 7.2606 & 0.0096 & 0.0090 & 2.8393 & 0.0248 & 0.0244 & 3.6473 & 0.5948 & 0.5021 & 17.5475 \\
NLinear & \textbf{0.1240} & 0.1060 & 10.3691 & 0.0555 & 0.0483 & 7.3240 & 0.0099 & 0.0092 & 2.8961 & 0.0173 & 0.0170 & 2.5386 & 0.5945 & 0.5021 & 17.5471 \\
TimesNet & 0.1756 & 0.1410 & 13.7828 & 0.1072 & 0.0920 & 12.8564 & 0.0576 & 0.0495 & 14.2102 & 0.1198 & 0.1057 & 15.7321 & 0.7049 & 0.5954 & 22.7788 \\
NBeats & 0.1670 & 0.1539 & 15.1466 & \underline{0.0528} & \underline{0.0453} & 7.6258 & 0.0093 & 0.0086 & 2.3634 & 0.0319 & 0.0249 & 3.8035 & 0.7117 & 0.5649 & 19.7326 \\
\bottomrule
\end{tabular}
}
\end{table}

\subsection{Experimental results}
Table~\ref{tab3} presents the main performance comparison of our proposed FAF framework against various time series forecasting methods on five real-world datasets. The results demonstrate that FAF achieves superior or competitive performance in nearly all datasets. As shown in the table~\ref{tab3}, FAF outperforms all baseline models on four out of the five datasets including Electricity, Walmart, CO$_2$, and GDP datasets. On the remaining dataset (Temperature dataset), our method maintains a strong performance, closely following the best-performing model without significant degradation. Notably, FAF achieves the best improvement on the CO$_2$ Emissions dataset, where the test set consist of 9 distinct tasks. This dataset poses a significant challenge due to its high inter-task variability. Our framework's design that decouples generalization features from task-specific features enables effective knowledge transfer while preserving local adaptability, leading to robust prediction accuracy in cold-start or few-shot time series forecasting scenarios.

\begin{figure}[!t]
\centering
\includegraphics[width=1.0\textwidth]{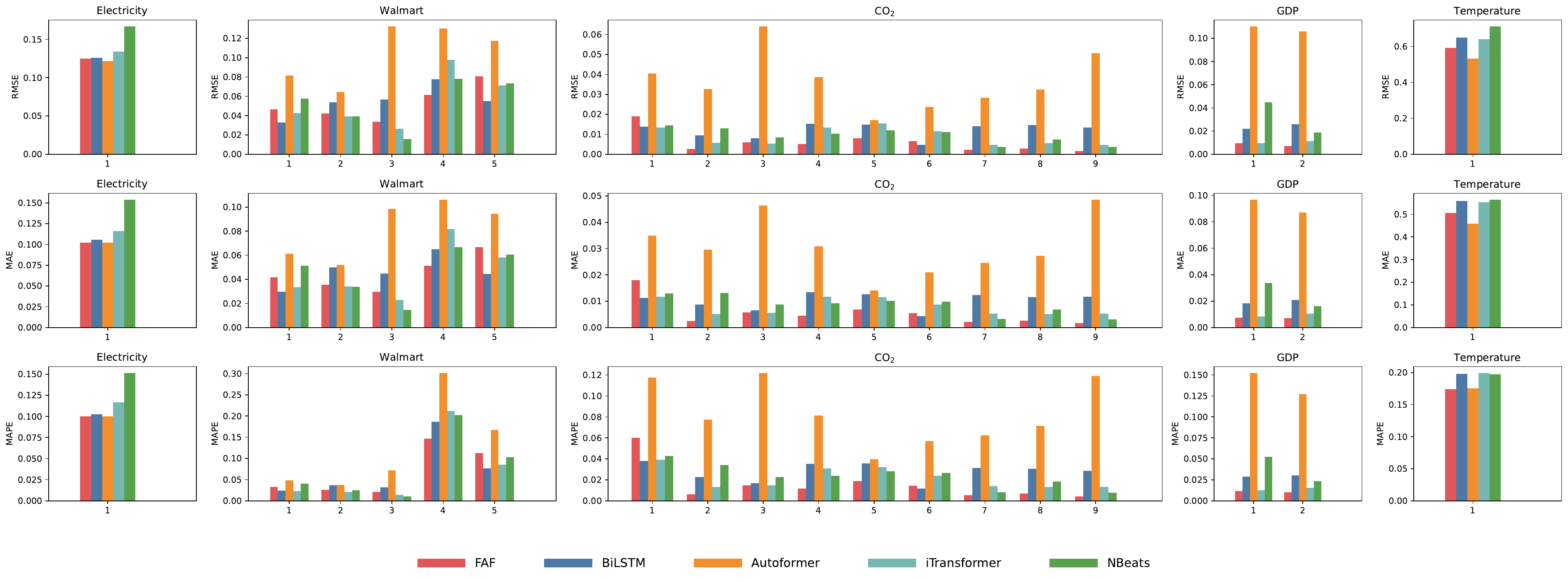}
\caption{Visualized performance comparison across individual test tasks.}
\label{fig3}
\end{figure}

To further investigate the performance of our model in few-shot time series forecasting scenario, we present per-task performance comparisons between FAF and selected baseline methods in Table~\ref{tab4}. This table focuses on four representative baseline methods including BiLSTM, Autoformer, iTransformer and NBeats.
As shown in Table~\ref{tab4} and the corresponding visualization (Figure~\ref{fig3}), FAF consistently outperforms these baselines across nearly all test tasks. For example, on the Walmart Sales dataset, FAF achieving the best or second-best performance in RMSE and MAE for most tasks, demonstrating superior adaptability to local variations with sparse historical data. Similarly, in the CO$_2$ Emissions dataset, which contains 9 distinct test tasks with high inter-task variability, FAF maintains stable performance and significantly reduces MAPE compared to other methods. Although some baselines such as Autoformer and iTransformer perform well on certain tasks, their performance tends to fluctuate substantially across different test tasks. In contrast, FAF exhibits consistent accuracy, indicating better generalization and more reliable adaptation capabilities in real-world few-shot time series forecasting scenarios.

\begin{table}[!t]
\centering
\caption{Per-task forecasting performance comparison of FAF versus selected baselines across all datasets.}
\label{tab4}
\resizebox{\textwidth}{!}{
\begin{tabular}{l c|ccc|ccc|ccc|ccc|ccc}
\hline
\multicolumn{2}{c|}{\multirow{2}{*}{Metric}}  & \multicolumn{3}{c|}{\textbf{FAF (ours)}} & \multicolumn{3}{c|}{BiLSTM} & \multicolumn{3}{c|}{Autoformer} & \multicolumn{3}{c|}{iTransformer} & \multicolumn{3}{c}{NBeats} \\
\cline{3-17}
& & RMSE & MAE & MAPE(\%) & RMSE & MAE & MAPE(\%) & RMSE & MAE & MAPE(\%) & RMSE & MAE & MAPE(\%) & RMSE & MAE & MAPE(\%) \\
\hline
{Electricity} & Test Task 1 & 0.1251 & 0.1021 & 10.0156 & 0.1258 & 0.1058 & 10.2257 & 0.1218 & 0.1024 & 10.0232 & 0.1340 & 0.1163 & 11.6500 & 0.1670 & 0.1539 & 15.1466 \\
\hline
\multirow{5}{*}{Walmart} & Test Task 1 & 0.0464 & 0.0417 & 3.2887 & 0.0326 & 0.0297 & 2.3551 & 0.0815 & 0.0612 & 4.8027 & 0.0427 & 0.0334 & 2.2875 & 0.0575 & 0.0512 & 4.0672 \\
& Test Task 2 & 0.0422 & 0.0355 & 2.6393 & 0.0538 & 0.0500 & 3.6706 & 0.0644 & 0.0520 & 3.7939 & 0.0394 & 0.0341 & 2.1555 & 0.0394 & 0.0338 & 2.4727 \\
& Test Task 3 & 0.0337 & 0.0298 & 2.1615 & 0.0568 & 0.0449 & 3.2287 & 0.1325 & 0.0987 & 7.1322 & 0.0266 & 0.0230 & 1.4205 & 0.0158 & 0.0146 & 1.0506 \\
& Test Task 4 & 0.0616 & 0.0512 & 14.6688 & 0.0779 & 0.0651 & 18.6687 & 0.1305 & 0.1062 & 30.1567 & 0.0977 & 0.0817 & 21.1950 & 0.0782 & 0.0666 & 20.2215 \\
& Test Task 5 & 0.0805 & 0.0667 & 11.2498 & 0.0551 & 0.0444 & 7.6439 & 0.1176 & 0.0945 & 16.7330 & 0.0715 & 0.0583 & 8.5445 & 0.0733 & 0.0605 & 10.3171 \\
\hline
\multirow{9}{*}{CO$_2$} & Test Task 1 & 0.0190 & 0.0180 & 6.0215 & 0.0139 & 0.0112 & 3.7970 & 0.0405 & 0.0350 & 11.7499 & 0.0134 & 0.0117 & 3.9057 & 0.0145 & 0.0129 & 4.2840 \\
& Test Task 2 & 0.0026 & 0.0024 & 0.6274 & 0.0095 & 0.0087 & 2.2607 & 0.0326 & 0.0296 & 7.7195 & 0.0058 & 0.0051 & 1.3286 & 0.0129 & 0.0132 & 3.4340 \\
& Test Task 3 & 0.0059 & 0.0057 & 1.5048 & 0.0081 & 0.0065 & 1.7008 & 0.0641 & 0.0463 & 12.1875 & 0.0053 & 0.0056 & 1.4842 & 0.0084 & 0.0087 & 2.2830 \\
& Test Task 4 & 0.0051 & 0.0045 & 1.1848 & 0.0152 & 0.0134 & 3.5405 & 0.0386 & 0.0308 & 8.1490 & 0.0134 & 0.0117 & 3.0795 & 0.0102 & 0.0092 & 2.4050 \\
& Test Task 5 & 0.0081 & 0.0068 & 1.8728 & 0.0149 & 0.0127 & 3.5576 & 0.0171 & 0.0141 & 3.9545 & 0.0155 & 0.0115 & 3.1956 & 0.0119 & 0.0101 & 2.8080 \\
& Test Task 6 & 0.0065 & 0.0054 & 1.4492 & 0.0047 & 0.0043 & 1.1607 & 0.0237 & 0.0210 & 5.6944 & 0.0115 & 0.0088 & 2.3741 & 0.0111 & 0.0099 & 2.6470 \\
& Test Task 7 & 0.0022 & 0.0022 & 0.5506 & 0.0140 & 0.0123 & 3.1290 & 0.0283 & 0.0246 & 6.2472 & 0.0048 & 0.0053 & 1.3909 & 0.0037 & 0.0033 & 0.8260 \\
& Test Task 8 & 0.0029 & 0.0027 & 0.7044 & 0.0146 & 0.0116 & 3.0543 & 0.0325 & 0.0273 & 7.1592 & 0.0056 & 0.0051 & 1.3300 & 0.0073 & 0.0069 & 1.8250 \\
& Test Task 9 & 0.0017 & 0.0017 & 0.4151 & 0.0133 & 0.0117 & 2.8714 & 0.0506 & 0.0486 & 11.9217 & 0.0048 & 0.0053 & 1.3412 & 0.0037 & 0.0031 & 0.7590 \\
\hline
\multirow{2}{*}{GDP} & Test Task 1 & 0.0095 & 0.0075 & 1.1760 & 0.0219 & 0.0185 & 2.9009 & 0.1104 & 0.0968 & 15.2267 & 0.0096 & 0.0083 & 1.2533 & 0.0449 & 0.0336 & 5.2666 \\
& Test Task 2 & 0.0071 & 0.0071 & 1.0233 & 0.0258 & 0.0208 & 3.0195 & 0.1060 & 0.0870 & 12.7157 & 0.0114 & 0.0105 & 1.5477 & 0.0188 & 0.0161 & 2.3403 \\
\hline
Temperature & Test Task 1 & 0.5917 & 0.5059 & 17.4309 & 0.6489 & 0.5583 & 19.8188 & 0.5332 & 0.4584 & 17.5299 & 0.6395 & 0.5530 & 19.9428 & 0.7117 & 0.5649 & 19.7326 \\
\hline
\end{tabular}
}
\end{table}

\begin{table}[!t]
\centering
\caption{Total and active parameter counts of FAF and selected baseline models.}
\label{tab5}
\tabcolsep 25pt %space between two columns. 用于调整列间距
\begin{tabular}{lcc}
\toprule
Model & Total Parameters & Active Parameters \\
\midrule
\textbf{FAF (Ours)} & \textbf{24,055} & \textbf{2,890} \\
NBeats & 51,078 & 51,078 \\
BiLSTM & 266,241 & 266,241 \\
iTransformer & 100,417 & 100,417 \\
Autoformer & 461,185 & 461,185 \\
\bottomrule
\end{tabular}
\end{table}

Table~\ref{tab5} summarizes the total and active parameter counts of FAF and selected strong baseline methods. Notably, FAF activates only a small subset (2,890 out of 24,055) of its parameters during inference, resulting in significantly reduced computational overhead compared to other models that utilize their full parameter sets for each task.This modular activation strategy not only enhances efficiency but also supports our design goal of balancing generalization and specialization. In contrast, methods like NBeats, BiLSTM, and Autoformer require full model capacity for every prediction, leading to higher memory consumption and slower adaptation speed, which is especially problematic in cold-start or few-shot time series forecasting scenarios where rapid deployment is important.

\subsection{Ablation study}
\textbf{Ablation study on modular design and functional regions count.} 
To analyze the contribution of each module in our framework, we conduct an ablation study by varying the number of functional regions within the Task-Specific Module and comparing performance with and without the Generalized Knowledge Module.

In this experiment, we denote $ S \in \{0, 1\} $ as whether the Generalized Knowledge Module is used (S=1) or not (S=0), and $ R \in \{2, 4, 8, 16\} $ as the number of activated functional regions in the Task-Specific Module. Table~\ref{tab6} presents the results across all five datasets.
As shown in Table~\ref{tab6}, removing the Generalized Knowledge Module (i.e., setting $ S = 0 $) leads to notable degradation in prediction accuracy, especially when the number of functional regions is small. This indicates that generalization features plays a crucial role in initializing task-specific adaptation in few-shot time series forecasting scenarios.
Moreover, increasing the number of functional regions improves model expressiveness and adaptability. The best results are achieved when both Generalized Knowledge Module and sufficient functional regions are available that validate our design choice of disentangling global trends from local dynamics.

\begin{table}[!t]
\centering
\caption{Ablation study on the Generalized Knowledge Module and functional regions count}
\label{tab6}
\resizebox{\textwidth}{!}{
\begin{tabular}{l|ccc|ccc|ccc|ccc|ccc}
\toprule
\multirow{2}{*}{Setting ($S$ / $R$)} & \multicolumn{3}{c|}{Electricity} & \multicolumn{3}{c|}{Walmart} & \multicolumn{3}{c|}{CO$_2$} & \multicolumn{3}{c|}{GDP} & \multicolumn{3}{c}{Temperature} \\
\cline{2-16}
& RMSE & MAE & MAPE (\%) & RMSE & MAE & MAPE (\%) & RMSE & MAE & MAPE (\%) & RMSE & MAE & MAPE (\%) & RMSE & MAE & MAPE (\%) \\
\midrule
FAF ($S=1$, $R=16$) & 0.1251 & 0.1021 & 10.0156 & 0.0529 & 0.0450 & 6.8016 & 0.0060 & 0.0055 & 1.5923 & 0.0083 & 0.0073 & 1.0997 & 0.5917 & 0.5059 & 17.4309 \\
FAF ($S=1$, $R=8$) & 0.1505 & 0.1222 & 12.1067 & 0.0436 & 0.0356 & 5.9380 & 0.0066 & 0.0060 & 1.6283 & 0.0649 & 0.0641 & 9.5154 & 0.6501 & 0.5468 & 19.1818 \\
FAF ($S=1$, $R=4$) & 0.1538 & 0.1283 & 12.8874 & 0.0554 & 0.0487 & 7.2679 & 0.0057 & 0.0050 & 1.3793 & 0.0722 & 0.0716 & 10.6404 & 0.6984 & 0.6183 & 21.3308 \\
FAF ($S=1$, $R=2$) & 0.1284 & 0.1077 & 10.6375 & 0.0742 & 0.0673 & 8.4392 & 0.0046 & 0.0038 & 1.0679 & 0.0684 & 0.0678 & 10.0814 & 0.6905 & 0.6038 & 21.0879 \\
\midrule
FAF ($S=0$, $R=16$) & 0.1417 & 0.1174 & 11.7700 & 0.0756 & 0.0629 & 9.7270 & 0.0166 & 0.0160 & 4.7218 & 0.0443 & 0.0436 & 6.6534 & 0.6047 & 0.5084 & 17.7921 \\
FAF ($S=0$, $R=8$) & 0.1498 & 0.1218 & 12.2028 & 0.0460 & 0.0377 & 6.1676 & 0.0041 & 0.0035 & 0.9755 & 0.0434 & 0.0429 & 6.5472 & 0.7409 & 0.6572 & 22.5537 \\
FAF ($S=0$, $R=4$) & 0.1474 & 0.1216 & 12.1955 & 0.0658 & 0.0591 & 7.9680 & 0.0046 & 0.0040 & 1.1390 & 0.0316 & 0.0308 & 4.6936 & 0.6406 & 0.5371 & 18.6842 \\
FAF ($S=0$, $R=2$) & 0.1253 & 0.1033 & 10.1161 & 0.0575 & 0.0502 & 6.9369 & 0.0064 & 0.0058 & 1.6229 & 0.0082 & 0.0077 & 1.1538 & 0.6439 & 0.5291 & 18.7132 \\
\bottomrule
\end{tabular}
}
\end{table}

\textbf{Impact of input and prediction sequence length.} 
To investigate how the length of historical input and prediction affects forecasting performance, we conduct an ablation study across four different settings, where both $ L $ (input length) and $ H $ (prediction length) are varied across test tasks. The configurations for each setting are summarized in Table~\ref{tab7}, while the corresponding results are reported in Table~\ref{tab8}.

\begin{figure}[!t]
\centering
\includegraphics[width=1.0\textwidth]{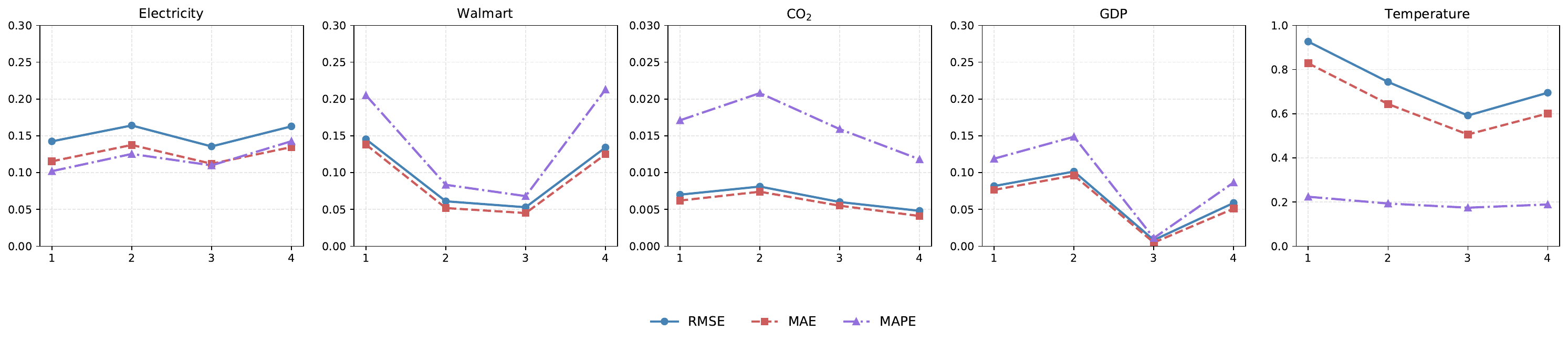}
\caption{Comparative visualization of FAF under various input-prediction length configurations.}
\label{fig4}
\end{figure}

As shown in Table~\ref{tab8} and visualized in Figure~\ref{fig4}, while longer historical sequences can improve accuracy to some extent, FAF demonstrates stable performance even under limited input lengths. Notably, setting 3 ($L=16, H=8$), which closely reflects real-world few-shot forecasting scenarios, achieves the best or near-best results on most datasets. This analysis further validates the robustness and effectiveness of FAF under data-scarce conditions. It demonstrates that our framework maintains strong predictive accuracy even with limited historical observations, aligning well with the few-shot time series forecasting objective.

\begin{table}[!t]
\centering
\caption{Input and prediction sequence length configurations.}
\label{tab7}
\tabcolsep 25pt %space between two columns. 用于调整列间距
\begin{tabular}{l|cccc}
\toprule
Setting & \textbf{1} & \textbf{2} & \textbf{3} & \textbf{4} \\
\midrule
Input Length $ L $ & 8 & 8 & 16 & 16 \\
Prediction Length $ H $ & 8 & 16 & 8 & 16 \\
\bottomrule
\end{tabular}
\end{table}

%%%%%%%%%%%%%%%%%%%%%%%%%%%%%%%%%%%%%%%%%%%%%%%%%%%%%%%
%%% 第五节
%%%%%%%%%%%%%%%%%%%%%%%%%%%%%%%%%%%%%%%%%%%%%%%%%%%%%%%
\section{Conclusion}
In this paper, we proposed FAF, a modular framework for few-shot time series forecasting that effectively disentangles generalization features from task-specific features. The framework consists of a Generalized Knowledge Module and multiple functional regions within the Task-Specific Module, which are dynamically selected by the Rank Module during prediction. This design enables adaptive features composition, allowing the model to balance generalization and specialization based on input sequence properties.
We evaluated FAF on five real-world datasets covering diverse domains such as electricity load
management, retail sales prediction, carbon emissions analysis, macroeconomic growth modeling, and
temperature monitoring. Experimental results demonstrate that FAF consistently outperforms state-of-the-art baseline methods in both overall and per-task performance. In particular, FAF achieves superior accuracy under short historical sequences, highlighting its robustness in cold-start or few-shot scenarios where conventional models struggle.
Ablation studies further validate the effectiveness of our method and the importance of dynamic module selection. Visualizations show how FAF activates only a subset of total parameters during inference, enhancing both computational efficiency and interpretability. These findings confirm that FAF mitigates feature conflicts through modular design and supports fast adaptation even with limited data. In the future, we aim to extend FAF to handle multivariate time series data by incorporating additional input modalities to better model complex dependencies in practical forecasting tasks. 

\begin{table}[!t]
\centering
\caption{FAF Performance under Different input-prediction Configurations}
\label{tab8}
\resizebox{\textwidth}{!}{
\begin{tabular}{l|ccc|ccc|ccc|ccc}
\toprule
Dataset & \multicolumn{3}{c|}{Setting 1} 
        & \multicolumn{3}{c|}{Setting 2}
        & \multicolumn{3}{c|}{Setting 3}
        & \multicolumn{3}{c}{Setting 4} \\
\cline{2-13}
& RMSE & MAE & MAPE (\%) & RMSE & MAE & MAPE (\%) & RMSE & MAE & MAPE (\%) & RMSE & MAE & MAPE (\%) \\
\midrule
Electricity & 0.1424 & 0.1153 & 10.19 & 0.1640 & 0.1375 & 12.53 & 0.1251 & 0.1021 & 10.02 & 0.1628 & 0.1346 & 14.26 \\
Walmart     & 0.1453 & 0.1380 & 20.50 & 0.0610 & 0.0518 & 8.34 & 0.0529 & 0.0450 & 6.80 & 0.1341 & 0.1248 & 21.29 \\
CO$_2$     & 0.0070 & 0.0062 & 1.71 & 0.0081 & 0.0074 & 2.08 & 0.0060 & 0.0055 & 1.59 & 0.0048 & 0.0041 & 1.18 \\
GDP         & 0.0817 & 0.0815 & 11.89 & 0.1014 & 0.1011 & 14.87 & 0.0083 & 0.0073 & 1.10 & 0.0588 & 0.0570 & 8.66 \\
Temperature & 0.9264 & 0.8283 & 22.39 & 0.7437 & 0.6437 & 19.32 & 0.5917 & 0.5059 & 17.43 & 0.6949 & 0.6011 & 18.89 \\
\bottomrule
\end{tabular}
}
\end{table}

\section*{Acknowledgment}
This work is supported by the National Natural Science Foundation of China (Grant Nos.62325602, 62036010).

%%%%%%%%%%%%%%%%%%%%%%%%%%%%%%%%%%%%%%%%%%%%%%%%%%%%%%%
%%% Appendix sections. 附录章节, 非必选
%%%%%%%%%%%%%%%%%%%%%%%%%%%%%%%%%%%%%%%%%%%%%%%%%%%%%%%
%\begin{appendix}
%\section{Name}

%\end{appendix}

\end{document}